%% file: root.tex
\documentclass[letterpaper, 10 pt, conference]{ieeeconf}  %

\IEEEoverridecommandlockouts                              %

\usepackage{amsmath}
\usepackage{algpseudocode}
\usepackage{algorithm}
\usepackage{pifont}
\newcommand{\cmark}{\textcolor{ForestGreen}{\ding{51}}}
\newcommand{\xmark}{\textcolor{BrickRed}{\ding{55}}}
\usepackage{listings}
\usepackage{color}
\usepackage[dvipsnames]{xcolor}
\usepackage{colortbl}
\usepackage{tabularx}
\usepackage[Export]{adjustbox}
\usepackage{subfig}
\usepackage{lipsum}
\usepackage{amsfonts}
\usepackage{hyperref}

\usepackage[normalem]{ulem}
\makeatletter
\providecommand{\theHALG@line}{}
\renewcommand{\theHALG@line}{\thealgorithm.\arabic{ALG@line}}
\makeatother
\usepackage{tikz}
\usepackage{pgfplots}
\usepackage{graphicx}
\usepackage{cleveref}
\usepackage{caption}
\newsavebox{\arrangebox}

\definecolor{reasonpurple}{RGB}{122,102,168}
\definecolor{bandcol}{HTML}{FFF7CC}
\newcommand{\band}[1]{\cellcolor{bandcol}#1}
\title{\LARGE \bf
ECHO: Embodied Camera Observations of Human Object Carrying
}

\author{Xuefei Sun$^{1}$, Lorin Achey$^{1}$, Kali Hamilton$^{1}$, Alberto Speranzon$^{2}$,\\
Gregory Grebe$^{2}$, Yonatan Bisk$^{3}$, Christoffer Heckman$^{1}$%
\thanks{$^{1}$X. Sun, L. Achey, K. Hamilton and C. Heckman are with the
Intelligent Robotics Laboratory, Department of Computer Science, University of
Colorado Boulder, Boulder, CO 80309, USA.
{\tt\small firstname.lastname@colorado.edu}}%
\thanks{$^{2}$A. Speranzon and G. S. Grebe are with Lockheed Martin, Advanced
Technology Laboratories, USA. {\tt\small \{alberto.speranzon,
gregory.s.grebe\}@lmco.com}}%
\thanks{$^{3}$Y. Bisk is with the Language Technologies Institute, Carnegie
Mellon University, Pittsburgh, PA 15213, USA.
{\tt\small ybisk@andrew.cmu.edu}}%
}

\begin{document}

\maketitle
\thispagestyle{empty}
\pagestyle{empty}

\input{sec/macros}

\input{sec/0_abstract}    
\input{sec/1_intro}

\input{sec/2_relatedworks}
\input{sec/3_preliminaries}
\input{sec/4_method}

\input{sec/5_results}

\input{sec/6_conclusion}

\section*{Acknowledgment}

This work was supported by Lockheed Martin, by the National Science Foundation
(NSF) under CAREER Award 2339328, and by the Army Research Laboratory (ARL)
Distributed and Collaborative Intelligent Systems and Technology (DCIST)
program under Cooperative Agreement W911NF-17-2-0181.
\bibliographystyle{IEEEtran}
\bibliography{main}

\end{document}

%% file: sec/macros.tex
\newcommand{\numScenes}{115}          %
\newcommand{\numFloors}{159}          %
\newcommand{\numPlans}{4,818}            %
\newcommand{\numEpisodes}{4,818}      %
\newcommand{\numAccepted}{3,805}      %
\newcommand{\acceptRate}{90\%}  %
\newcommand{\numObjects}{198}        %
\newcommand{\poolSize}{198}                   %
\newcommand{\numSources}{five}                %
\newcommand{\meanCapLen}{22.2}    %
\newcommand{\numSurfInstances}{1,704} %
\newcommand{\numRoomPairs}{157}    %
\newcommand{\numAnnotators}{4}  %
\newcommand{\medCoverage}{1.00}  %
\newcommand{\meanScanFrames}{563} %
\newcommand{\framesEnc}{30}                   %
\newcommand{\numEnc}{3}                       %
\newcommand{\numMultiStorey}{42}
\newcommand{\meanRoomsFloor}{5.6}
\newcommand{\meanSurfFloor}{36}
\newcommand{\meanRoute}{12.3}
\newcommand{\medRoute}{10.9}
\newcommand{\crossRoom}{96\%}
\newcommand{\tierTypicalPct}{36\%}
\newcommand{\tierPlausiblePct}{30\%}
\newcommand{\tierAtypicalPct}{24\%}
\newcommand{\tierOffPriorPct}{10\%}
\newcommand{\nonCanonicalPct}{64\%}
\newcommand{\ctxVocab}{4{,}112}
\newcommand{\ctxMTLD}{72.8}
\newcommand{\capVocab}{196}
\newcommand{\meanCtxLen}{19.0}
\newcommand{\ctxUniquePct}{99.8\%}
\newcommand{\leakNeitherPct}{65\%}
\newcommand{\leakBothPct}{11\%}
\newcommand{\modalSurfPct}{31\%}
\newcommand{\modalRoomPct}{33\%}
\newcommand{\condEntropy}{2.65}
\newcommand{\distinctSurfPerObj}{7.9}
\newcommand{\ambGeTwoPct}{86\%}
\newcommand{\ambMedian}{4}
\newcommand{\ambRoomGeTwoPct}{52\%}
\newcommand{\numTriples}{2,886}
\newcommand{\numSurfClasses}{23}
\newcommand{\epsPerInstance}{1.9}
\newcommand{\levelExplicitPct}{37\%}
\newcommand{\levelSingleHopPct}{34\%}
\newcommand{\levelMultiHopPct}{29\%}
\newcommand{\wupSibling}{0.88}
\newcommand{\wupBowl}{0.82}
\newcommand{\cosPitcher}{0.50}
\newcommand{\cosBowl}{0.54}
\newcommand{\cosBox}{0.15}
\newcommand{\wnFloorPct}{70\%}
\newcommand{\capEditPct}{12\%}
\newcommand{\ctxRewritePct}{78\%}
\newcommand{\capTemplates}{639}

%% file: sec/0_abstract.tex
\begin{abstract}
Embodied and assistive agents must do more than recognize objects: they must
reason about \emph{where an object belongs} given the layout of an environment
and the habits of the people who live in it. However, there has been limited progress on problems like this. One explanation is the lack of a dedicated benchmark or dataset for defining and evaluating the problem. Existing RGB-D scan datasets reconstruct static rooms but contain no human activity, while
human-object-interaction datasets capture motion without a navigable, fully
reconstructed scene or a ground-truth notion of an object's natural
destination. We introduce \emph{contextual object placement} as a benchmark task  for predicting an object's destination during an observed object carrying episode. To support this task, we present Embodied Camera observations of
Human Object carrying (\emph{ECHO}), a large-scale synthetic dataset that
pairs dense RGB-D scans of indoor scenes with recordings of an embodied human
carrying everyday objects to context-appropriate destinations. ECHO is the first publicly available dataset to combine reconstructed scenes, human activity, %
natural language, and contextual-placement annotations. ECHO comprises
\numAccepted{} human-annotated episodes across \numFloors{} floors of
\numScenes{} HM3D scenes, carrying \numObjects{} distinct objects. Each floor includes a complete RGB-D scan with human-annotated room labels and a surface
list. Each episode provides synchronized RGB-D encounter clips, 6-DoF camera,
human, and object trajectories, start and destination surfaces, an action
caption and a human-written \emph{context}: a single sentence describing the inhabitant's routine that implies the destination without explicitly naming it. We evaluate contextual object placement using input-masked probes and an end-to-end baseline. Results show that no single input modality is sufficient, highlighting the need to jointly reason over scene structure, human activity, and contextual knowledge.
\end{abstract}

%% file: sec/1_intro.tex
\section{Introduction}
\label{sec:intro}

\begin{figure}[t]
  \centering
  \includegraphics[width=\linewidth]{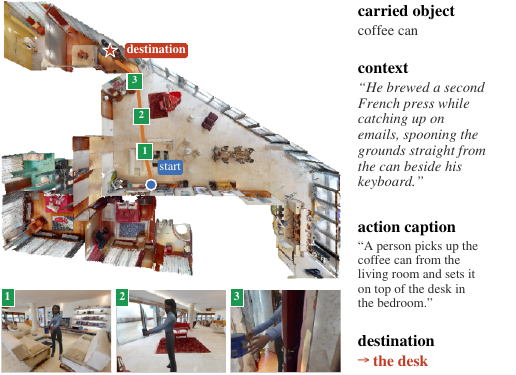}
  \caption{\textbf{One ECHO episode.} The information available to the agent includes the RGB-D floor scan and three encounter clips along the carrying route (left). The context provides a hint about the destination surface (right). The dataset additionally provides the carried object name, action caption, and ground-truth destination.}
  \label{fig:teaser}
\end{figure}

A service robot that operates in a home alongside a person must repeatedly answer a
question that object recognition alone cannot resolve: \emph{where does this object
belong?} The answer determines whether a tidying robot returns a mug to the drying rack its owner uses every morning or to a cupboard nobody opens, whether a robotic assistant can anticipate where a person carrying a book is headed, and whether it can act without being explicitly instructed.

The difficulty is that object semantics alone does not determine the answer. A home offers many plausible locations for a given object to be placed or, when carried, many possible destinations; a book could reasonably belong on a shelf, a desk, a coffee table, or a nightstand. General knowledge about where books are usually seen cannot distinguish among these alternatives. What breaks the tie is the inhabitant's routine: the book belongs on \emph{their} nightstand because they read before sleeping. That preference is almost never stated explicitly. Instead, it must be inferred from indirect evidence about the inhabitant's routine, what we call \emph{context}, and grounded in the home's geometry, where multiple surfaces may be plausible destinations for the object.%

We pose this inference
as \emph{contextual object placement}
(Sec.~\ref{sec:prelims}). Given a dense RGB-D scan of the environment, an embodied observation of a person mid-carry, and person-specific contextual information, the agent must predict the surface on which the carry will end. The destination depends on two kinds of evidence. Object semantics together with the person's habits indicate \emph{what kind of surface} is appropriate for the carried object, while the observed motion and scene geometry indicate \emph{which instance} of that surface the person is approaching. The observation covers only part of the
carry, so the destination remains a prediction: context and object type narrows the set of plausible surfaces, and the partially observed trajectory grounds that prediction to a specific surface among many candidates in the home. Despite its importance, no existing benchmark %
jointly provides the scene geometry, human behavior, and contextual information needed to study this problem. %

We supply the data for this task with ECHO, a large-scale synthetic dataset collected in Habitat~\cite{savva2019habitat} over HM3D~\cite{ramakrishnan2021hm3d} scenes. In each episode, an embodied human avatar carries an everyday object while a robot agent records both a scene sweep and the carrying interaction (Fig.~\ref{fig:teaser}). Expert annotators wrote or revised the context for every episode and reviewed each episode through a two-stage quality-control process. The resulting dataset captures placements that cannot be explained by common object-location associations alone and often require distinguishing among multiple plausible destination surfaces. These properties make ECHO a challenging benchmark for studying contextual object placement.

Our contributions are:
\begin{itemize}
  \item \textbf{The ECHO dataset:} \numAccepted{} human-annotated carry
episodes across \numFloors{} floors of \numScenes{} HM3D scenes, carrying
\numObjects{} objects from \numSources{} corpora. Each floor includes an RGB-D
scan with human-annotated room labels, a list of candidate surfaces, and
spawned objects. Each episode provides synchronized RGB-D encounter clips;
6-DoF camera, human, and object trajectories; start and destination
surfaces; an action caption; a human-written context; and placement and
context-difficulty labels (Table~\ref{tab:contents}).
  \item \textbf{The contextual object-placement task:} an application-grounded
  formalization (Sec.~\ref{sec:prelims}) with an evaluation protocol that
  scores the destination location in terms of room, as well as surface category and  instance.
  \item \textbf{Baseline evaluation:} input-masked probes show 
  that no single input channel resolves the destination, and an end-to-end
  baseline shows that imperfect perception costs accuracy mainly in grounding
  the prediction to a surface instance, not in choosing the kind of surface.
\end{itemize}

%% file: sec/2_relatedworks.tex
\section{Related Works}
\label{sec:RW}

\textbf{3D scene-understanding datasets.}
ScanNet~\cite{dai2017scannet}, Matterport3D~\cite{Matterport3D},
Gibson~\cite{xia2018gibson}, and HM3D~\cite{ramakrishnan2021hm3d,yadav2023hm3dsem} provide
large-scale RGB-D reconstructions of real interiors, but capture only static
geometry, appearance, and semantics. They do not capture dynamic changes in
the environment, including changes involving movable objects or human
activities. ECHO builds on HM3D and draws its carried objects from
ReplicaCAD~\cite{szot2021habitat2}, YCB~\cite{calli2015ycb}, and the
GSO~\cite{downs2022gso}, ABO~\cite{collins2022abo}, and
HSSD~\cite{khanna2023hssd} subsets of the OVMM~\cite{yenamandra2023homerobot} object set. It extends the
static scene representation with human object-carrying episodes, contextual
descriptions, and destination annotations.

\textbf{Embodied-AI benchmarks.}
Simulators such as
Habitat~\cite{savva2019habitat,szot2021habitat2,puig2023habitat3},
AI2-THOR/ProcTHOR~\cite{kolve2017ai2thor,deitke2022procthor}, and
iGibson/BEHAVIOR~\cite{li2021igibson,srivastava2022behavior} support a range
of placement-related tasks. Object-goal navigation~\cite{batra2020objectnav}
specifies a target category and
rearrangement~\cite{batra2020rearrangement,weihs2021rearrangement} a target
goal state. Housekeep~\cite{kant2022housekeep}, TIDEE~\cite{sarch2022tidee}, and PARSEC~\cite{ramachandruni2025parsec} instead infer where objects belong from commonsense or user preference, but the system chooses the placement itself, drawing on category-level knowledge learned from training scenes or crowdsourced arrangements without observing a human. In ECHO, the
signal comes from an observed human: models watch the carry in dense
RGB-D and predict the destination implied by the person's routine, rather than
an explicitly specified goal or a population-level prior.

\textbf{Human activity and goal inference.}
Existing work on human activity and goal inference includes datasets that
capture human manipulation, motion, and task execution in indoor environments. HOI4D~\cite{liu2022hoi4d}, BEHAVE~\cite{bhatnagar2022behave},
CORE4D~\cite{liu2025core4d}, Ego-Exo4D~\cite{grauman2024egoexo4d}, and
EPIC-KITCHENS~\cite{damen2022epic100} capture real manipulation but without navigable scene scans; BEHAVIOR-1K~\cite{li2023behavior1k}
simulates a thousand activities toward defined goal states; and
HUMANISE~\cite{wang2022humanise} aligns motion capture with ScanNet rooms
for language-conditioned motion generation. As summarized in
Table~\ref{tab:compare}, these datasets provide subsets of the signals
required for contextual object placement, but none pair human activity in
navigable environments with annotations of a carried object's
context-dependent destination. In language-guided embodied tasks, REVERIE~\cite{qi2020reverie} and
SIF~\cite{min2024sif} provide instructions that name the goal or follow
fixed templates, whereas ECHO provides lexically diverse free-form contextual descriptions from which the destination must be inferred (Table~\ref{tab:langdiv}). Most closely related, LP$^2$~\cite{gorlo2024lp2} forecasts human trajectories by prompting an LLM over a privileged symbolic scene graph, and notes that habit cues could sharpen its predictions. ECHO instead grounds human-object carrying behavior in contextual habits, combining human motion, scene structure, language, and destination annotations at scale. To our knowledge, ECHO is the first dataset to combine reconstructed scenes, human activities, natural language, and contextual-placement annotations(Table~\ref{tab:compare}).

\begin{table}[t]
  \centering
  \caption{Comparison of ECHO with related scene, embodied-AI, and
human-activity datasets. Scan = a navigable RGB-D reconstruction of the full
environment released with the dataset; Place.\ = supervision of an object's destination placement.
\cmark$^{\dagger}$ denotes a partial property.}
  \label{tab:compare}
  \small
  \setlength{\tabcolsep}{3.2pt}
  \resizebox{\linewidth}{!}{%
  \begin{tabular}{@{}lcccccr@{}}
    \hline
    Dataset & Scan & Hum.\ traj. & Obj.\ traj. & Lang. & Place. & Scale \\
    \hline
    ScanNet~\cite{dai2017scannet}        & \cmark & \xmark & \xmark & \xmark & \xmark & 1{,}513 scans \\
    Matterport3D~\cite{Matterport3D}     & \cmark & \xmark & \xmark & \xmark & \xmark & 90 buildings \\
    HM3D~\cite{ramakrishnan2021hm3d}     & \cmark & \xmark & \xmark & \xmark & \xmark & 1{,}000 scenes \\
    ScanRefer~\cite{chen2020scanrefer}   & \cmark & \xmark & \xmark & \cmark & \xmark & 51.6K descr. \\
    REVERIE~\cite{qi2020reverie}         & \cmark & \xmark & \xmark & \cmark & \xmark & 21.7K instr. \\
    Rearrange.~\cite{batra2020rearrangement} & \xmark & \xmark & \xmark & \xmark & \cmark & --- \\
    LP$^2$~\cite{gorlo2024lp2}           & \xmark & \cmark & \xmark & \cmark & \xmark & 2 envs \\
    SIF~\cite{min2024sif}                & \xmark & \cmark & \xmark & \cmark & \cmark & 480 tasks \\
    \hline
    HOI4D~\cite{liu2022hoi4d}            & \xmark & \cmark$^{\dagger}$ & \cmark & \xmark & \xmark & 4.0K seq \\
    BEHAVE~\cite{bhatnagar2022behave}    & \xmark & \cmark & \cmark & \xmark & \xmark & 321 seq \\
    CORE4D~\cite{liu2025core4d}          & \xmark & \cmark & \cmark & \xmark & \xmark & 1K real seq \\
    Ego-Exo4D~\cite{grauman2024egoexo4d} & \xmark & \cmark & \xmark & \cmark & \xmark & 1{,}286 h \\
    EPIC-KITCHENS~\cite{damen2022epic100} & \xmark & \cmark$^{\dagger}$ & \cmark$^{\dagger}$ & \cmark & \xmark & 90K actions \\
    BEHAVIOR-1K~\cite{li2023behavior1k}  & \xmark & \xmark & \cmark & \cmark$^{\dagger}$ & \cmark & 1K activities \\
    \hline
    \textbf{ECHO (ours)}                 & \cmark & \cmark & \cmark & \cmark & \cmark & \numAccepted{} eps \\
    \hline
  \end{tabular}}
\end{table}

%% file: sec/3_preliminaries.tex
\section{Problem Definition}
\label{sec:prelims}

\subsection{Task definition}
\label{subsec:taskdef}
We define contextual object placement as the task of placing an object based on contextual cues. Abstractly such tasks may present with a range of contextual information. On one end of the spectrum, the context is empty and placement relies only on commonsense knowledge; at the other, the goal location is stated directly through an instruction. ECHO focuses on the intermediate case, where the intended destination must be inferred from trajectory observations and contextual information that do not explicitly state goal locations. %
Overall, ECHO covers a wide subset of these tasks, specifically focused on domestic object placement problems.

Formally, the agent observes a scene scan $\mathcal{O}_{\text{scan}}$
(Sec.~\ref{subsec:contents}) 
consisting of posed RGB-D frames collected throughout the environment. The scan defines a set of
candidate placement surfaces $\mathcal{P} = \{p_m\}$, where each surface
$p_m$ has a room label $r(p_m)$ and location $\mathbf{x}(p_m)$.  The agent then encounters a person carrying an object through a second observation stream, \(\mathcal{O}_{\text{enc}}\), collected along the agent's own camera trajectory and covering only part of the carrying activity. The agent also receives a human-written contextual description $c$ associated with the episode. The agent's contextual object placement problem may be formulated as:

\begin{center}
\setlength{\fboxsep}{6pt}%
\fbox{\parbox{0.92\columnwidth}{\small Given the sweep
$\mathcal{O}_{\text{scan}}$, the encounter $\mathcal{O}_{\text{enc}}$, and a
context $c$, predict the destination surface
$\hat{p}_{\text{end}} \in \mathcal{P}$ on which the carried object will be placed.}}
\end{center}

Because the encounter happened before the placement, predicting
$\hat{p}_{\text{end}}$ is genuine \emph{prediction} from two complementary
cues: the perceived motion, which constrains where the carrier is headed, and the
context, which enables down- selecting candidate destination surfaces.%

\subsection{Evaluation protocol}
\label{subsec:eval}
We evaluate performance through destination prediction. To distinguish between partially correct predictions, we measure accuracy at three levels of granularity: (i)~room accuracy (Acc$_{\text{room}}$), which measures whether \(r(\hat{p}_{\text{end}})\) matches the annotated destination room; (ii)~category accuracy (Acc$_{\text{cat}}$), which measures whether the predicted surface category matches the annotated destination category; and (iii)~instance accuracy (Acc$_{\text{inst}}$), which measures whether \(\hat{p}_{\text{end}}\) corresponds to the exact annotated destination surface instance. The gap between category and instance accuracy reflects the benchmark's \emph{geometry gap}: cases in which a method identifies the correct type of surface but fails to localize the correct physical instance.

As a secondary, geometry-aware measure we report placement error, the
Euclidean distance $\lVert \mathbf{x}(\hat{p}_{\text{end}}) -
\mathbf{x}(p_{\text{end}}) \rVert_2$ between the predicted and annotated
placement locations, which credits near-misses that land on the wrong surface instance but in the
correct region of the environment.

%% file: sec/4_method.tex
\section{The ECHO Dataset}
\label{sec:method}

We first describe the released dataset contents, then the collection
pipeline, the placement-difficulty controls used during generation, and the
resulting dataset statistics. %

\subsection{The ECHO dataset}
\label{subsec:contents}

\begin{table}[t]
  \centering
  \caption{What ECHO releases, per floor and per episode. \emph{Role} marks
  whether an artifact is an input under the task definition of
  Sec.~\ref{sec:prelims} or ground-truth supervision (GT) only.}
  \label{tab:contents}
  \footnotesize
  \setlength{\tabcolsep}{3pt}
  \renewcommand{\arraystretch}{1.12}
  \begin{tabular}{@{}>{\raggedright\arraybackslash}p{1.95cm}>{\raggedright\arraybackslash}p{5.6cm}@{\hspace{6pt}}l@{}}
    \hline
    Artifact & Content & Role \\
    \hline
    \multicolumn{3}{@{}l}{\rule{0pt}{2.6ex}\textbf{Per floor} --- \numFloors{} floors of \numScenes{} HM3D scenes} \\
    $\mathcal{O}_{\text{scan}}$ & RGB-D frames with poses and intrinsics;
      $640{\times}480$, $90^\circ$ HFOV, $30$\,fps; \meanScanFrames{} frames
      on average & input \\
    Room labels & human-assigned room type for every region & GT \\
    Surface list $\mathcal{P}$ & scene objects with class, center, extent,
      and room label & GT \\
    Objects & spawned objects with label, start surface, and frozen start
      pose & GT \\
    \hline
    \multicolumn{3}{@{}l}{\rule{0pt}{2.6ex}\textbf{Per episode} --- \numAccepted{} accepted episodes, \numObjects{} objects} \\
    $\mathcal{O}_{\text{enc}}$ & \numEnc{} clips of \framesEnc{} RGB-D frames,
      with camera, human, and object 6-DoF per frame & input \\
    Context $c$  & human-written sentence of the inhabitant's routine & input \\
    Action caption & templated sentence naming object, origin, destination & GT \\
    Human track & full-carry 6-DoF trajectory of the avatar & GT \\
    Carried object & label and start surface & GT \\
    Destination & surface instance, class, room, placement position & GT \\
    Placement tier & typical / plausible / atypical destination & GT \\
    Context level & explicit / implicit single-hop / implicit multi-hop & GT \\
    Split & scene-disjoint train / test (80/20)
     & --- \\
    \hline
  \end{tabular}
\end{table}

Under the task definition of
Sec.~\ref{sec:prelims} an agent receives two
types of input images: a scene scan, which provides observations of the environment, and the encounter clips of each episode, which capture a person carrying an object. 
Additionally, we provide human annotations beyond those available in HM3D~\cite{ramakrishnan2021hm3d,yadav2023hm3dsem}. Room labels are assigned by annotators because HM3D defines room boundaries but does not provide semantic room names. The surface list corresponds to the candidate placement set $\mathcal{P}$ introduced in Sec.~\ref{sec:prelims}, containing every valid destination surface on the floor.

Each episode contains two language annotations with distinct roles. The action caption explicitly names the carried object, origin, and destination in templated form, whereas the context is a free-form description of the inhabitant's routine that implies the destination without stating it directly. We also provide two difficulty labels: the placement tier, which measures how much the destination departs from the object's typical placement prior (Sec.~\ref{subsec:tiers}), and the context reasoning difficulty level, which measures the number of reasoning steps required to infer the destination from the provided context (Sec.~\ref{subsec:analysis}).

\subsection{Collection pipeline}
\label{subsec:pipeline}

\textbf{Scenes and objects.} Scenes are drawn from
HM3D and processed per floor, since one building
reconstruction may span several floors with disconnected navigable areas;
floors with very limited room or surface diversity, such as an empty basement
level, are discarded by hand. The pool of carriable
objects combines assets from \numSources{} sources: the hand-authored
ReplicaCAD~\cite{szot2021habitat2} household assets, the
YCB benchmark~\cite{calli2015ycb}, and curated subsets of Google Scanned
Objects (GSO)~\cite{downs2022gso}, Amazon Berkeley Objects
(ABO)~\cite{collins2022abo}, and HSSD~\cite{khanna2023hssd}. For each floor, a subset of the \poolSize{} curated objects is spawned based
on the floor's size and room count. The carrying human model is sampled from a
set of eight rigged avatars \cite{hoangminh2022humanmodel}. %
Candidate placement surfaces are extracted from
HM3D semantics and annotated room labels. Together, these surfaces and the
spawned objects define the scene state used throughout the episode generation
process.

\textbf{Episode planning.} Episode generation is divided into a floor-level
 and an episode-level. The floor-level determines where each
object begins, while the episode-level determines where it will be
placed after the carry. This separation allows multiple episodes to share the same initial
scene configuration while varying the destination.

For each floor, we use Qwen3-8B to assign spawned objects to initial surfaces using a
scene summary and the placement priors described in
Sec.~\ref{subsec:tiers}. Objects are then physically instantiated and settled
with Bullet %
onto the available free space of the assigned surface, after
which their poses are frozen. As a result, all episodes on this floor, as well
as the scene scan, share the same object configuration. Starting from this shared configuration, episode-level planning generates three episodes for each object and fixed start surface by assigning distinct destination surfaces according to the target distribution of placement-difficulty tiers. Once the object and destination are fixed, the human
trajectory, action caption, and contextual description are generated and
grounded to the selected geometry.

\textbf{Human review.} %
A context must imply the object's destination through routine without naming the destination surface or room which makes the
task challenging for untrained annotators\cite{mauceri2019sun}. We therefore hired three expert annotators and paired them with four draft-generating LLMs: Claude Sonnet 5, Claude Opus 5, Claude Haiku 4.5, and DeepSeek V4 Flash. This resulted in twelve annotator--model pairings and reduced reliance on any single annotator or model.

Every episode undergoes two stages of expert review.
First, annotators verify the floor-level scene configuration, inspecting
object initial poses from eye-level and top-down renderings and revising
implausible placements when necessary. Second, annotators review each episode's trajectory, caption, and context draft, ensuring that the context provides a precise description of the intended destination without naming the surface directly, and correcting any mislabeled or inconsistent annotations.

\textbf{Two-phase recording.} Each floor is first recorded through a scene
scan, in which an autonomous explorer traverses all reachable areas and
collects RGB-D observations of the environment. In addition to the recorded observations, we provide the object meshes and
poses used in each scene. Together with the corresponding HM3D assets, these
annotations allow users to reconstruct the start state and render arbitrary
observations in Habitat. For each episode, the carried object is placed at its fixed start pose and an
avatar carries it along the planned route. Encounter clips are recorded from
the agent's viewpoint along the trajectory, providing
partial observations of the carrying activity while keeping the carried object
visible to the camera.
\begin{figure}[t]
  \centering
  \includegraphics[width=0.8\linewidth]{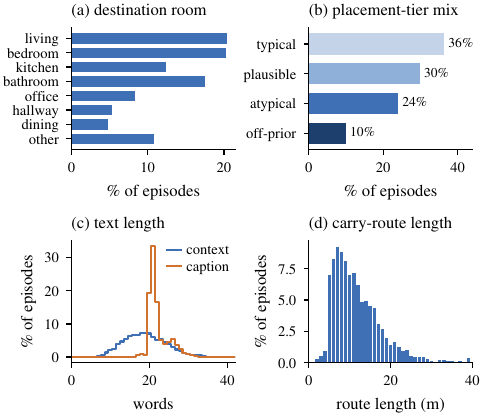}
  \caption{Dataset distributions: (a)~destination rooms,
  (b)~placement-tier mix, (c)~caption and context length, (d)~carry-route
  length.}
  \label{fig:stats}
\end{figure}
\subsection{Controlling placement difficulty}
\label{subsec:tiers}
Contextual object placement is most challenging when the destination cannot
be inferred from common object-placement patterns alone. A model that relies
only on object semantics can often ignore the context and default to the most
typical destination: a mug goes to the kitchen, a book goes to a shelf.
ECHO therefore controls placement difficulty during episode generation by
varying how strongly the destination agrees with these priors. %
For each carried object, we estimate object--room and object--surface priors
from large-scale annotated indoor datasets. The object--room prior is derived
from HM3D and Matterport3D while the object--surface prior is
estimated from ScanNet. To
improve coverage for sparsely observed objects, counts are pooled across
functionally similar categories using ConceptNet
Numberbatch~\cite{speer2017conceptnet}. Both
priors are restricted to the rooms and surface classes present on the current
floor and renormalized.

Using both priors, each episode is assigned to one of three
placement tiers: \emph{typical}, where the destination is the object's most
common location; \emph{plausible}, where the destination is not the
most likely choice; and \emph{atypical}, where the destination is rare and
must be justified by the context. To ensure coverage across a range of placement difficulty, episodes are
generated to target an equal distribution across the three tiers. The final
dataset contains approximately
$\tierTypicalPct/\tierPlausiblePct/\tierAtypicalPct$ typical, plausible, and
atypical placements, respectively, with \tierOffPriorPct{}\ off-prior
destinations; deviations from the target arise when particular placement
tiers are infeasible in scenes that lack suitable destination surfaces
(Fig.~\ref{fig:stats}b). %

\begin{figure*}[t]
  \centering
  \includegraphics[width=0.9\linewidth]{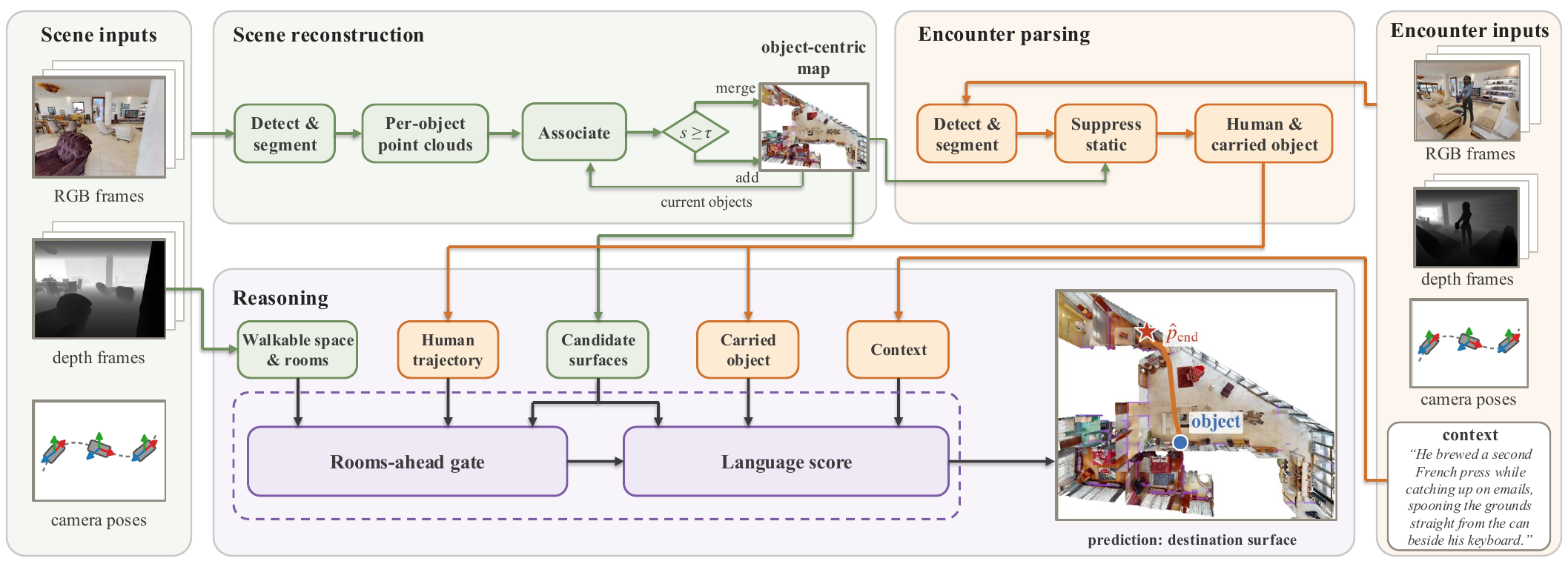}
  \caption{\textbf{ECHO end-to-end baseline.} Scene-scan observations
(\textcolor{green!50!black}{green}) are reconstructed into an object-centric
map from which candidate surfaces are extracted, while the depth frames
provide the walkable space and its rooms. Encounter observations
(\textcolor{orange}{orange}) recover the human trajectory and the carried
object. The reasoning stage (\textcolor{reasonpurple}{purple}) applies a
rooms-ahead gate from the observed trajectory on the walkable space, followed
by a language score conditioned on the carried object and the context, to
predict the destination surface.%
}
  \label{fig:pipeline}
\end{figure*}

\subsection{Dataset analysis}
\label{subsec:analysis}

\textbf{Spatial structure.} Each floor offers on average \meanRoomsFloor{}
room types. \crossRoom{} of episodes
cross a room boundary, with carry routes averaging \meanRoute{}\,m
(Fig.~\ref{fig:stats}d). Destinations spread over \numSurfClasses{} surface classes and
\numRoomPairs{} start$\to$end room pairs (Fig.~\ref{fig:stats}a), resulting in
\numSurfInstances{} distinct surface instances at \epsPerInstance{} episodes per instance; Across the dataset, we have \numTriples{} distinct combinations of carried object,
destination surface class, and destination room, so surface reuse adds
contrast rather than repetition.

\begin{figure}[t]
  \centering
  \includegraphics[width=0.8\linewidth]{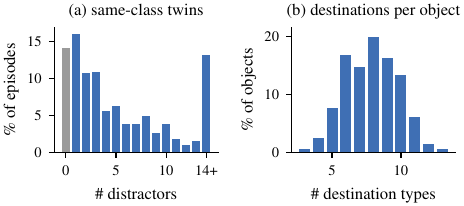}
  \caption{Why contextual object placement is difficult. (a)~Most
  destinations have several same-class competitors on their floor, so
  class-level reasoning cannot identify the instance. (b)~Destination
  classes reached per object: no object is tied to a single class, motivating the use of context.}
  \label{fig:ambiguity}
\end{figure}

\textbf{Destinations are context and instance dependent.} Object identity alone does not determine where an object belongs. Among objects appearing in at least five episodes, each reaches \distinctSurfPerObj{} destination-surface classes on average, corresponding to a conditional entropy of \condEntropy{} bits, or about six equally likely destination choices (Fig.~\ref{fig:ambiguity}b). Even after identifying the correct category, multiple surface instances often remain: among non-floor destinations, \ambGeTwoPct{} share their floor with another surface of the same class, with a median of \ambMedian{} competitors, while \ambRoomGeTwoPct{} have a same-class twin in the destination room (Fig.~\ref{fig:ambiguity}a). Thus, identifying the destination requires both contextual information and geometric grounding, motivating the task and the distinction between Acc\textsubscript{cat} and Acc\textsubscript{inst} in Sec.~\ref{subsec:eval}.

\textbf{Language.} ECHO provides two types of language annotations: context and caption. Table~\ref{tab:langdiv} compares their lexical diversity with language-paired 3D datasets using a common
word tokenizer and size-normalized statistics. Vocabulary and distinct-$n$ are
computed on equal-size 1{,}500-token subsamples averaged over 20 random draws,
while MATTR-50, HD-D, and MTLD %
 use standard length-robust measures. The contexts are highly diverse: \ctxUniquePct{} of contexts are
globally unique and lead all diversity measures, with an MTLD of \ctxMTLD{},
at least $2.5\times$ that of Nr3D, REVERIE, and ScanRefer. %
\begin{table}[t]
  \centering
  \caption{Language diversity of the free-form layer, measured uniformly across
  the released corpora as described in Sec.~\ref{subsec:analysis}. 
}
  \label{tab:langdiv}
  \small
  \setlength{\tabcolsep}{4pt}
  \resizebox{\linewidth}{!}{%
  \begin{tabular}{@{}lrrrrrr@{}}
    \hline
    Corpus & Vocab$\uparrow$ & D-2$\uparrow$ & D-3$\uparrow$ & MATTR-50$\uparrow$ & HD-D$\uparrow$ & MTLD$\uparrow$ \\
    \hline
    SIF~\cite{min2024sif}               &  67 & .09 & .14 & .46 & .48 & 21.8 \\
    Sr3D~\cite{achlioptas2020referit3d} & 116 & .16 & .34 & .50 & .52 & 13.5 \\
    REVERIE~\cite{qi2020reverie}        & 223 & .45 & .74 & .62 & .66 & 26.2 \\
    ScanRefer~\cite{chen2020scanrefer}  & 266 & .51 & .75 & .61 & .65 & 21.8 \\
    Nr3D~\cite{achlioptas2020referit3d} & 298 & .57 & .83 & .66 & .69 & 28.5 \\
    LP$^2$~\cite{gorlo2024lp2}          & 388 & .69 & .88 & .70 & .78 & 27.4 \\
    \hline
    ECHO captions                        & 102 & .13 & .21 & .48 & .49 & 26.5 \\
    \textbf{ECHO contexts}              & \textbf{551} & \textbf{.82} & \textbf{.96} & \textbf{.80} & \textbf{.81} & \textbf{72.8} \\
    \hline
  \end{tabular}}
\end{table}

\textbf{Context reasoning difficulty.} Predicting a destination from context can
require different amounts of reasoning. We therefore assign each context one
of three reasoning-difficulty levels. A context is \emph{explicit} if it
directly mentions the destination room or surface, \emph{implicit single-hop} if
it names a landmark within $2$m of the destination or a concept that is one
ConceptNet~\cite{speer2017conceptnet} relation away from the destination, and
\emph{implicit multi-hop} if its stated concepts reach the destination only
through two or more relations, or through no relation found by the search.
For example, ``she puts the cup on the sink'' is explicit, ``she brushes her
teeth'' is single-hop, and ``she gets ready for work'' is multi-hop. The three
levels account for 37\%, 34\%, and 28\% of contexts, respectively.

%% file: sec/5_results.tex
\section{Benchmark, Baseline, and Experiments}
\label{sec:experiments}

We instantiate the contextual object-placement benchmark of
Sec.~\ref{sec:prelims} on ECHO and evaluate methods using the room,
surface category, and instance metrics defined there. We first analyze the benchmark
with input-masked probes that isolate individual sources of information,
then report an end-to-end baseline that replaces oracle inputs with
perceived scene and trajectory information.

Episodes are split scene-disjointly into train and test sets at an 80/20 ratio.
All floors of the same HM3D scene belong to the same split.

\subsection{Input-masked probes}
\label{subsec:probes}
\textbf{Probes definition: }We use input-masked probes to measure how much
information each input provides. Every
probe receives the ground-truth surface list, including the class and room label of each
candidate object, together with one or more additional inputs: train-split
destination frequencies (\textsc{Prior-freq}), the carried-object label
(\textsc{Object}), the context (\textsc{Context}), the carrier trajectory
during the encounter (\textsc{Trajectory}), or their combinations
(\textsc{Context+Object}, \textsc{Traj+Object},
\textsc{Traj+Context}, and \textsc{Traj+Context+Object}). Single-input probes
measure the information available from each channel alone, while combined
probes measure the information gained by adding one channel to another.

\textsc{Prior-freq} uses destination frequencies from the training
split. \textsc{Object} and \textsc{Context} ask an LLM to score each \emph{(class, room)} group
given the carried object or context, respectively, while
\textsc{Context+Object} provides both inputs together. \textsc{Trajectory} fits a heading to the tracked human positions during the encounter and scores candidate surfaces by their bearing from that heading. \textsc{Traj+Context} instead uses a geodesic
\emph{rooms-ahead} gate on the floor's walkable grid. Groups along the carrier's
route through the floor's door topology receive higher scores, while receding
or unreachable groups are downweighted. \textsc{Traj+Object} and
\textsc{Traj+Context+Object} use the same geometry gate with their respective
language scores.

\begin{table}[t]
  \centering
  \caption{Input-masked probe accuracy on the 400-episode test sample at early,
mid, and late encounters. Results are reported at the room, category, and
instance levels; bold denotes the best result per column, and shading marks
results within $0.05$ of the best. }
  \label{tab:probes}
  \small
  \setlength{\tabcolsep}{3pt}
  \resizebox{\linewidth}{!}{%
  \begin{tabular}{@{}lccccccccc@{}}
    \hline
    & \multicolumn{3}{c}{early} & \multicolumn{3}{c}{mid} & \multicolumn{3}{c}{late} \\
    Probe & room & cat. & inst. & room & cat. & inst. & room & cat. & inst. \\
    \hline
    \textsc{Prior-freq} & 0.235 & 0.065 & 0.047 & 0.233 & 0.063 & 0.046 & 0.234 & 0.065 & 0.048 \\
    \textsc{Object} & 0.200 & 0.111 & 0.034 & 0.199 & 0.108 & 0.033 & 0.200 & 0.110 & 0.034 \\
    \textsc{Context} & 0.592 & \band{0.192} & \band{0.113} & 0.591 & \band{0.190} & 0.110 & 0.587 & 0.191 & 0.111 \\
    \textsc{Trajectory} & 0.333 & 0.010 & 0.000 & 0.491 & 0.023 & 0.012 & 0.667 & 0.049 & 0.052 \\
    \hline
    \textsc{Context+Object} & 0.576 & \band{0.204} & \band{0.115} & 0.575 & \band{0.202} & \band{0.112} & 0.571 & 0.200 & 0.112 \\
    \textsc{Traj+Object} & 0.296 & 0.123 & 0.056 & 0.450 & 0.133 & 0.062 & 0.650 & 0.215 & 0.180 \\
    \textsc{Traj+Context} & \band{\textbf{0.650}} & \band{0.207} & \band{0.127} & \band{\textbf{0.712}} & \band{0.201} & \band{0.147} & \band{\textbf{0.774}} & \band{0.249} & \band{0.189} \\
    \textsc{Traj+Context+Object} & \band{0.627} & \band{\textbf{0.226}} & \band{\textbf{0.135}} & \band{0.692} & \band{\textbf{0.237}} & \band{\textbf{0.161}} & \band{0.755} & \band{\textbf{0.290}} & \band{\textbf{0.234}} \\
    \hline
  \end{tabular}}
\end{table}

\begin{figure*}[t]
  \centering
  \includegraphics[width=0.9\linewidth]{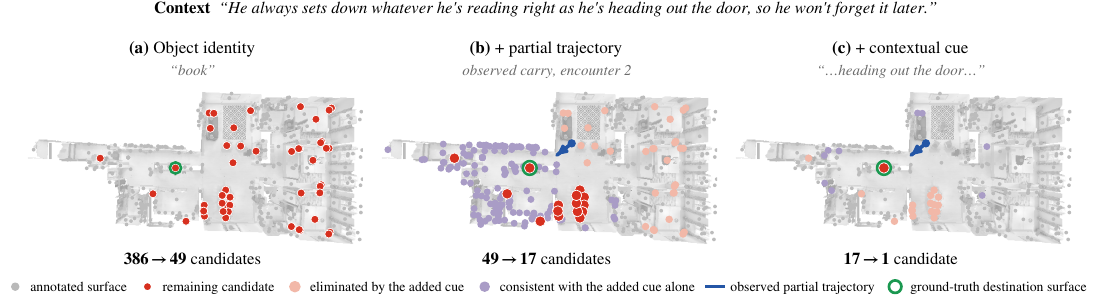}
  \caption{\textbf{Each cue alone is ambiguous; together they identify the
  surface.} In this test episode at the mid-route encounter, object identity narrows 386 candidate surfaces to 49, the observed trajectory reduces those to 17, and the contextual cue reduces the candidate set to a single surface.}
  \label{fig:narrowing}
\end{figure*}

\textbf{Probe results: }Table~\ref{tab:probes} reports probe accuracy on a 400-episode sample of the
test split at early, mid, and late encounters. Context is the strongest
static signal, whereas trajectory becomes increasingly informative as the
encounter progresses along the route. Combining both sources of information
produces the strongest performance throughout the carry.

The gap between room and instance accuracy remains large, indicating that
identifying the destination type is substantially easier than grounding the
correct surface instance. Adding the carried-object label contributes little
beyond context alone, suggesting that contextual information already captures
much of the object's semantic signal.

All LLM-backed probes use DeepSeek V4 Flash. We compared six different LLMs 
on the \textsc{Object} and \textsc{Context+Object}
probes over the same sample: adding context improved every model, and the
models did not differ significantly from one another.
\begin{table}[t]
  \centering
  \caption{Probe accuracy by context-difficulty level on the 400-episode
test sample at the mid-route encounter.} %
  \label{tab:levels}
  \small
  \setlength{\tabcolsep}{3pt}
  \resizebox{\linewidth}{!}{%
  \begin{tabular}{@{}lrccccccccc@{}}
    \hline
    & & \multicolumn{3}{c}{\textsc{Context}} & \multicolumn{3}{c}{\textsc{Traj+Context}} & \multicolumn{3}{c}{\textsc{Prior-freq}} \\
    Level & $n$ & room & cat. & inst. & room & cat. & inst. & room & cat. & inst. \\
    \hline
    explicit            & 148 & \textbf{0.694} & \textbf{0.261} & 0.129 & \textbf{0.813} & \textbf{0.272} & \textbf{0.172} & \textbf{0.253} & 0.047 & 0.031 \\
    implicit single-hop & 138 & 0.549 & 0.206 & \textbf{0.132} & 0.649 & 0.184 & 0.154 & 0.234 & \textbf{0.072} & 0.051 \\
    implicit multi-hop  & 113 & 0.506 & 0.077 & 0.065 & 0.655 & 0.130 & 0.110 & 0.206 & 0.071 & \textbf{0.058} \\
    \hline
  \end{tabular}}
\end{table}

Table~\ref{tab:levels} shows that context difficulty affects prediction as
intended. Context-only accuracy drops from explicit to implicit contexts,
especially at the instance level, while the frequency prior remains low across
all levels. The two implicit levels have similar room accuracy but differ more
clearly at the class and instance levels. Adding trajectory recovers part of
the loss, particularly for multi-hop contexts. Figure~\ref{fig:narrowing}
illustrates one episode at the mid-route encounter.

\subsection{End-to-end baseline}
\label{subsec:baseline}

Our end-to-end baseline (Fig.~\ref{fig:pipeline}) implements the
\textsc{Traj+Context} probe using a scene reconstructed from RGB-D
observations rather than ground-truth geometry. The scene is built using an
open-vocabulary object mapping pipeline following ConceptGraphs~\cite{conceptgraphs}.
YOLO-World~\cite{yoloworld} detections are segmented with SAM~\cite{sam},
back-projected using depth, and fused across frames using geometric overlap
and DINOv2~\cite{dinov2} features. The resulting map provides candidate
surfaces and their classes, while the depth observations provide the
walkable space used for the trajectory-based geodesic gate. Rooms are
inferred from the reconstructed walkable space and assigned to candidate
surfaces. During the encounter, detections are projected into the same world
frame, and moving evidence is linked across frames by nearest-centroid
association to obtain person and carried-object tracks. %
Table~\ref{tab:baseline} repeats the eight probes of
Table~\ref{tab:probes} with every oracle replaced by the baseline's
perception, on the same 400-episode test sample and encounters. Comparing a
cell against its oracle counterpart isolates the cost of perceiving that
channel.

\textbf{Trajectory is the easiest signal to recover.} The tracker detects the
person in $98.6\%$ and the carried object in $89.0\%$ of encounters, with
median horizontal errors of $0.13$ and $0.05$\,m, respectively.
Consequently, the perceived \textsc{Trajectory} probe matches the oracle's
placement error and stays within $0.03$ of its room accuracy at every
encounter point.

\textbf{Surface reconstruction is the primary bottleneck.} Although $97\%$ of
reconstructed points lie within $10$\,cm of the HM3D mesh, only $57\%$ of
annotated placement surfaces receive any reconstructed geometry and
only $23\%$ receive a class that we would accept compared to the ground-truth
label. Some of this gap may arise because surfaces that are visible in
the RGB-D observations, particularly those farther from the camera, receive
low detection confidence and are filtered out by the ConceptGraph pipeline.
As a result, the perceived \textsc{Traj+Context} probe retains only $0.047$ of
the oracle's $0.189$ late-encounter instance accuracy and $0.098$ of its
$0.249$ surface-category accuracy.

\textbf{Semantic errors add further loss.} The perceived room layer correctly
identifies the room for $55\%$ of matched surfaces, while object detection
fails to recover a track in $11\%$ of encounters. Overall, the largest gap
between oracle and perceived performance comes from reconstructing and predicting surface labels rather than estimating motion,
highlighting scene understanding as a key challenge for contextual object
placement.
Overall, the benchmark remains challenging when considering realistic perception pipelines. Closing this gap requires more reliable scene understanding as well as
contextual reasoning. ECHO provides a benchmark for measuring both.

\begin{table}[t]
  \centering
  \caption{End-to-end baseline on the same 400-episode test sample and
encounters as Table~\ref{tab:probes}. Each probe is evaluated using
perceived inputs from the pipeline in Fig.~\ref{fig:pipeline}: reconstructed
candidate surfaces and room labels, detected object classes, and tracked
human and object trajectories. Columns report destination room, category,
and instance accuracy. Reconstructed predictions are matched and grounded
to annotated objects on the same floor.}
  \label{tab:baseline}
  \small
  \setlength{\tabcolsep}{3pt}
  \resizebox{\linewidth}{!}{%
  \begin{tabular}{@{}lccccccccc@{}}
    \hline
    & \multicolumn{3}{c}{early} & \multicolumn{3}{c}{mid} & \multicolumn{3}{c}{late} \\
    Head & room & cat. & inst. & room & cat. & inst. & room & cat. & inst. \\
    \hline
    \textsc{Prior-freq} & 0.233 & 0.028 & 0.000 & 0.233 & 0.028 & 0.000 & 0.236 & 0.028 & 0.000 \\
    \textsc{Object} & 0.206 & 0.042 & 0.012 & 0.201 & 0.038 & 0.010 & 0.237 & 0.032 & 0.008 \\
    \textsc{Context} & 0.341 & 0.070 & 0.025 & 0.342 & 0.070 & 0.025 & 0.347 & 0.072 & 0.026 \\
    \textsc{Trajectory} & 0.358 & 0.027 & 0.006 & 0.475 & 0.034 & 0.015 & \textbf{0.641} & 0.062 & 0.043 \\
    \hline
    \textsc{Context+Object} & 0.318 & 0.058 & 0.024 & 0.335 & 0.059 & 0.024 & 0.345 & 0.055 & 0.020 \\
    \textsc{Traj+Object} & 0.280 & 0.056 & 0.016 & 0.388 & 0.072 & 0.020 & 0.574 & 0.074 & 0.035 \\
    \textsc{Traj+Context} & \textbf{0.406} & \textbf{0.084} & 0.033 & 0.480 & \textbf{0.087} & \textbf{0.035} & 0.534 & \textbf{0.098} & 0.047 \\
    \textsc{Traj+Context+Object} & 0.392 & 0.069 & \textbf{0.034} & \textbf{0.484} & 0.076 & 0.034 & 0.551 & 0.094 & \textbf{0.054} \\
    \hline
  \end{tabular}}
\end{table}

%% file: sec/6_conclusion.tex
\section{Conclusions and Future Work}
\label{sec:conclusion}

We introduced ECHO, a large-scale synthetic dataset for contextual object
placement, pairing reconstructed indoor scenes with embodied observations of
humans carrying everyday objects to context-appropriate destinations. ECHO
provides synchronized RGB-D observations, motion trajectories, scene
structure, and natural-language context, together with explicit placement and
context difficulty labels. We formalized contextual object placement as a
benchmark task and evaluated it with input-masked probes and an end-to-end
baseline. Our experiments show that no single input channel is sufficient:
context provides a strong static signal, trajectory becomes increasingly
informative during the carry, and their combination performs best. Under
realistic perception, however, reconstructing and predicting candidate surface labels is a larger source of error than recovering motion,
highlighting the importance of both scene understanding and contextual
reasoning.

\textbf{Future work.} We will evaluate learned contextual-placement methods
and human performance, providing both algorithmic baselines and a human
reference. We plan to expand ECHO to more scenes and objects, support
multi-object carrying, and extend single-sentence contexts toward persistent
inhabitant routines and behaviors. Finally, we will study transfer from ECHO
to real RGB-D observations and physical embodied agents, narrowing the gap
between synthetic benchmark evaluation and deployment.